\documentclass[10pt,twocolumn,letterpaper]{article}

\usepackage{cvpr}
\usepackage{times}
\usepackage{epsfig}
\usepackage{graphicx}
\usepackage{amsmath}
\usepackage{amssymb}
\usepackage{subfigure}
\usepackage{multirow}
\usepackage{booktabs}

\usepackage[breaklinks=true,bookmarks=false]{hyperref}

\cvprfinalcopy 

\ifcvprfinal\fi
\begin{document}

\title{Deep Spectral Clustering using Dual Autoencoder Network}

\author{Xu Yang\textsuperscript{1} , Cheng Deng\textsuperscript{1}\thanks{Corresponding author.} , Feng Zheng\textsuperscript{2} , Junchi Yan\textsuperscript{3} , Wei Liu\textsuperscript{4}$^{*}$\\
\textsuperscript{1}School of Electronic Engineering, Xidian University, Xian 710071, China\\
\textsuperscript{2}Department of Computer Science and Engineering, Southern University of Science and Technology\\
\textsuperscript{3}Department of CSE, and MoE Key Lab of Artificial Intelligence, Shanghai Jiao Tong University \\
\textsuperscript{4}Tencent AI Lab, Shenzhen, China \\
{\tt\small \{xuyang.xd, chdeng.xd\}@gmail.com, zhengf@sustc.edu.cn, }
\\ {\tt\small yanjunchi@sjtu.edu.cn, wl2223@columbia.edu}
}

\maketitle

\begin{abstract}
The clustering methods have recently absorbed even-increasing attention in learning and vision. Deep clustering combines embedding and clustering together to obtain optimal embedding subspace for clustering, which can be more effective compared with conventional clustering methods. In this paper, we propose a joint learning framework for discriminative embedding and spectral clustering. We first devise a dual autoencoder network, which enforces the reconstruction constraint for the latent representations and their noisy versions, to embed the inputs into a latent space for clustering. As such the learned latent representations can be more robust to noise. Then the mutual information estimation is utilized to provide more discriminative information from the inputs. Furthermore, a deep spectral clustering method is applied to embed the latent representations into the eigenspace and subsequently clusters them, which can fully exploit the relationship between inputs to achieve optimal clustering results. Experimental results on benchmark datasets show that our method can significantly outperform state-of-the-art clustering approaches.
\end{abstract}

\section{Introduction}
As an important task in unsupervised learning~\cite{yang2018semantic,deng2019unsupervised,li2019coupled,li2016scalable} and vision communities~\cite{yu2018incremental}, clustering~\cite{hoi2010semi} has been widely used in image segmentation~\cite{shi2000normalized}, image categorization~\cite{yang2019weightreg,yi2015efficient}, and digital media analysis~\cite{an2012robust}. The goal of clustering is to find a partition in order to keep similar data points in the same cluster while dissimilar ones in different clusters. In recent years, many clustering methods have been proposed, such as $K$-means clustering~\cite{macqueen1967some}, spectral clustering~\cite{ng2002spectral,Yang2018NewL1,jiang2012transfer}, and non-negative matrix factorization clustering~\cite{xu2003document}, among which $K$-means and spectral clustering are two well-known conventional algorithms that are applicable to a wide range of various tasks. However, these shallow clustering methods depend on low-level features such as raw pixels, SIFT~\cite{ng2003sift} or HOG~\cite{dalal2005histograms} of the inputs. Their distance metrics are only exploited to describe local relationships in data space, and have limitation to represent the latent dependencies among the inputs~\cite{chen2018deep}.

This paper presents a novel deep learning based unsupervised clustering approach. Deep clustering, which integrates embedding and clustering processes to obtain optimal embedding subspace for clustering, can be more effective than shallow clustering methods. The main reason is that the deep clustering methods can effectively model the distribution of the inputs and capture the non-linear property, being more suitable to real-world clustering scenarios.

\begin{figure}
 \centering
  \subfigure[Raw data]{
  \centering
  \label{fig:subfigs:tezheng1} 
  \includegraphics[height=2.65cm,width=2.4cm,]{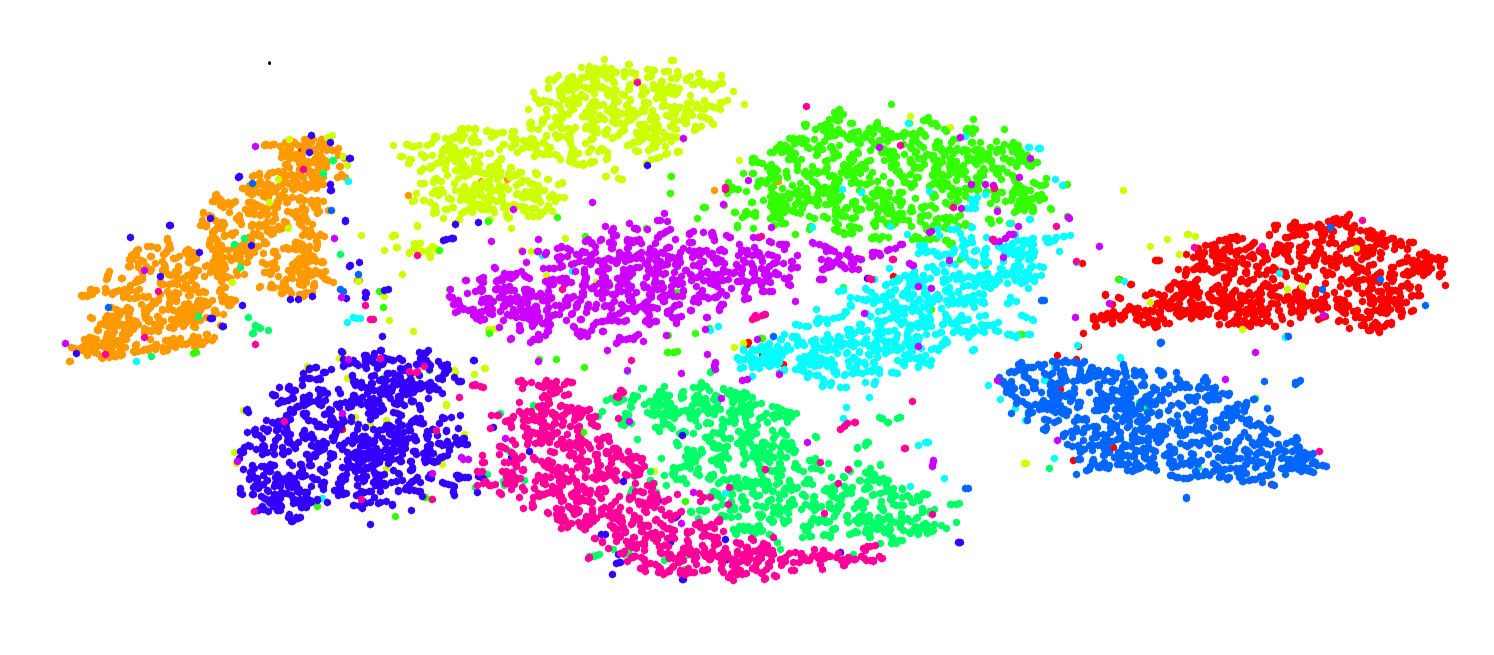}}
  \hspace{.01in}
   \subfigure[ConvAE]{
  \centering
  \label{fig:subfigs:tezheng2} 
  \includegraphics[height=2.65cm,width=2.4cm]{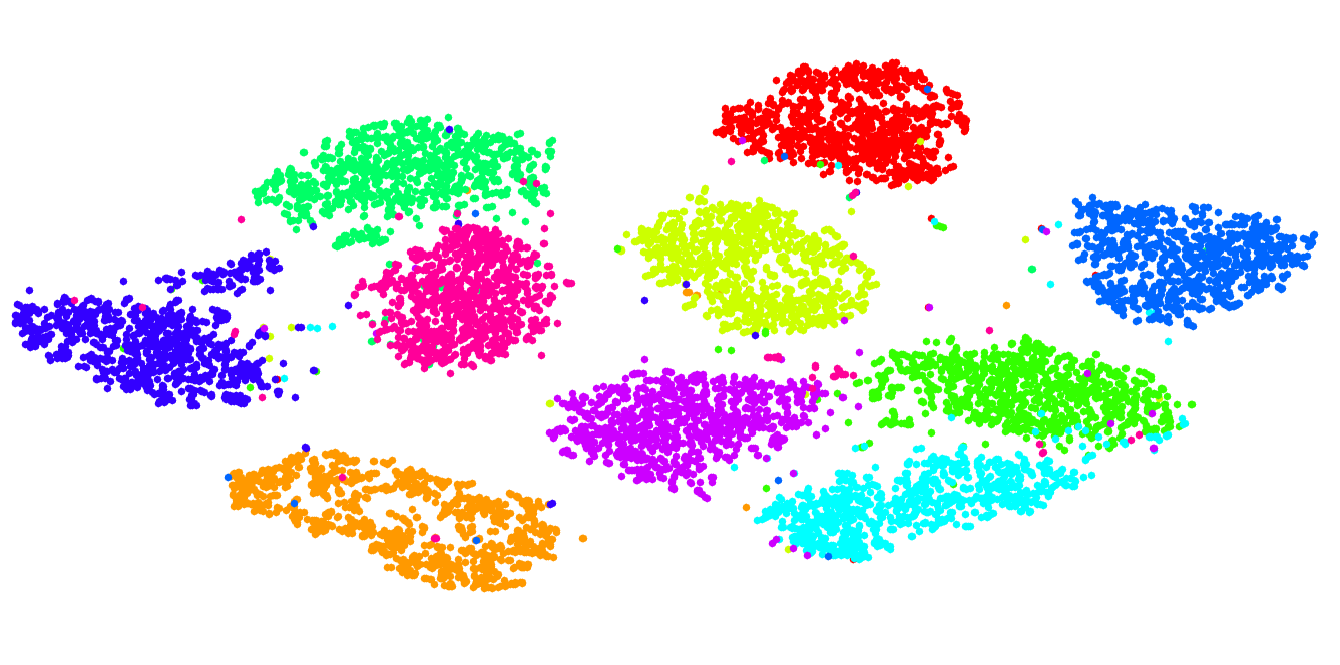}}
  \hspace{.01in}
   \subfigure[Our method]{
  \centering
  \label{fig:subfigs:tezheng3} 
  \includegraphics[height=2.65cm,width=2.4cm]{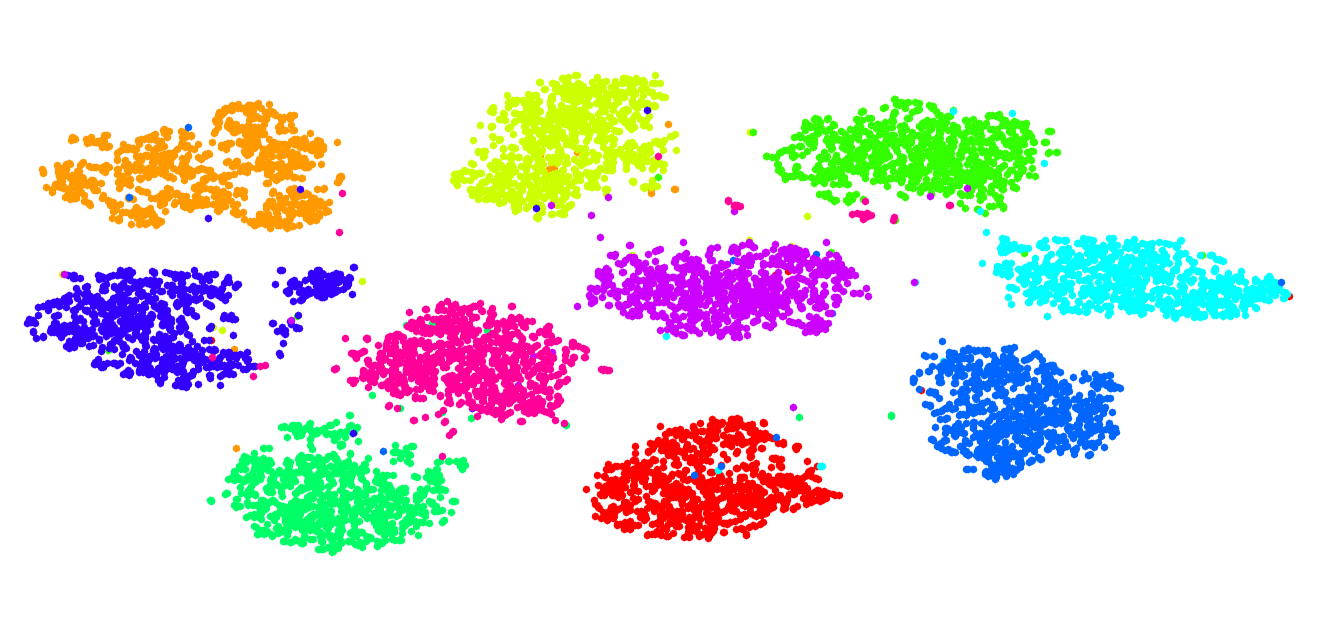}}
   \caption{Visualizing the discriminative embedding capability on MNIST-\emph{test} with $t$-SNE algorithm. (a): the space of raw data, (b): data points in the latent subspace of convolution autoencoder; (c): data points in the latent subspace of the proposed autoencoder network. Our method can provide a more discriminative embedding subspace.}
   \label{fig: differentfeatures} 
 \end{figure}
Recently, many clustering methods are promoted by deep generative approaches, such as autoencoder network~\cite{masci2011stacked}. The popularity of the autoencoder network lies in its powerful ability to capture high dimensional probability distributions of the inputs without supervised information. The encoder model projects the inputs into the latent space, and adopts an explicit approximation of maximum likelihood to estimate the distribution diversity between the latent representations and the inputs. Simultaneously, the decoder model reconstructs the latent representations to ensure the output maintaining all of the details in the inputs~\cite{tzoreff2018deep}. Almost all existing deep clustering methods endeavor to minimize the reconstruction loss. The hope is making the latent representations more discriminative which directly determines the clustering quality. However, in fact, the discriminative ability of the latent representations has no substantial connection with the reconstruction loss, causing the performance gap that is to be bridged in this paper.

We propose a novel dual autoencoder network for deep spectral clustering. First, a dual autoencoder, which enforces the reconstruction constraint for the latent representations and their noisy versions, is utilized to establish the relationships between the inputs and their latent representations. Such a mechanism is performed to make the latent representations more robust. In addition, we adopt the mutual information estimation to reserve discriminative information from the inputs to an extreme. In this way, the decoder can be viewed as a discriminator to determine whether the latent representations are discriminative. Fig.~\ref{fig: differentfeatures} demonstrates the performance of our proposed autoencoder network by comparing different data representations on MNIST-\emph{test} data points. Obviously, our method can provide more discriminative embedding subspace than the convolution autoencoder network. Furthermore, deep spectral clustering is harnessed to embed the latent representations into the eigenspace, which followed by clustering. This procedure can exploit the relationships between the data points effectively and obtain the optimal results. The proposed dual autoencoder network and deep spectral clustering network are jointly optimized.

The main contributions of this paper are in three-folds:

\begin{itemize}
\item We propose a novel dual autoencoder network for generating discriminative and robust latent representations, which is trained with the mutual information estimation and different reconstruction results.
\item We present a joint learning framework to embed the inputs into a discriminative latent space with a dual autoencoder and assign them to the ideal distribution by a deep spectral clustering model simultaneously.
\item Empirical experiments demonstrate that our method outperforms state-of-the-art methods over the five benchmark datasets, including both traditional and deep network-based models.
\end{itemize}
\section{Related Work}
Recently, a number of deep learning-based clustering methods are proposed. Deep Embedding Clustering~\cite{xie2016unsupervised} (DEC) adopts a fully connected stacked autoencoder network in order to learn the latent representations by minimizing the reconstruction loss in the pre-training phase. The objective function applied to the clustering phase is the Kullback Leibler ($KL$) divergence between the soft assignments of clustering modelled by a $t$-distribution. And then, a $K$-means loss is adopted at the clustering phase to train a fully connected autoencoder network~\cite{yang2016towards}, which is a joint approach of dimensionality reduction and $K$-means clustering. In addition, Gaussian Mixture Variational Autoencoder (GMVAE)~\cite{dilokthanakul2016deep} shows that minimum information constraint can be utilized to mitigate the effect of over-regularization in VAEs and provides an unsupervised clustering within the VAE framework considering a Gaussian mixture as a prior distribution. Discriminatively Boosted Clustering~\cite{li2018discriminatively}, a fully convolutional network with layer-wised batch normalization, adopts the same objective function as DEC and uses a boosting factor to the relatively train a stacked autoencoder.
\begin{figure*}[!ht]
\centering
\includegraphics[height=6cm,width=0.9\textwidth]{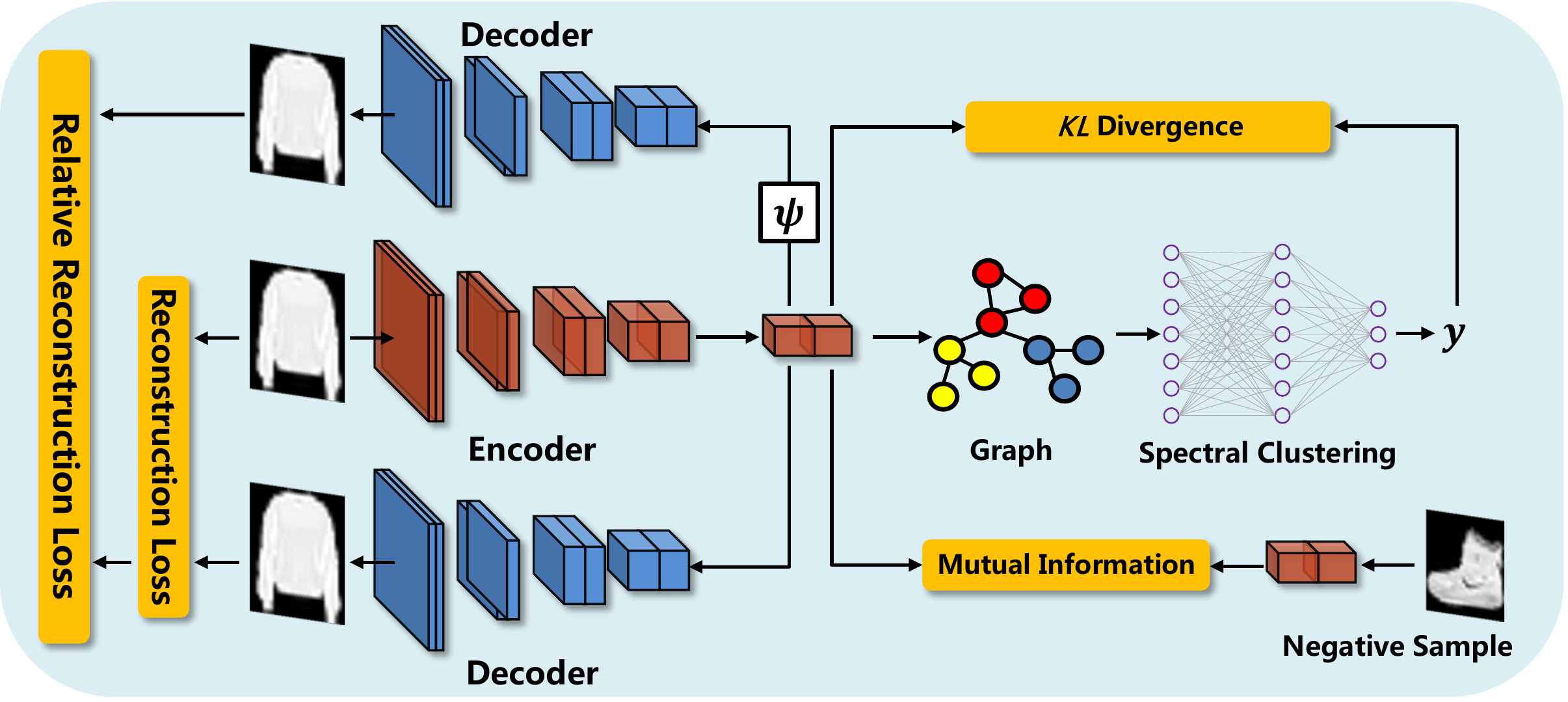}\\
\caption{Illustration of the overall architecture. We first pre-train a dual autoencoder to embed the inputs into a latent space, and reconstruction results are obtained by the latent representations and their noise versions based on the noisy-transformer $\psi$. The mutual information calculated with negative sampling estimation is utilized to learn the discriminative information from inputs. Then, we assign the latent representations to the ideal clusters by a deep spectral clustering model, and jointly optimize the dual autoencoder and spectral clustering network simultaneously.}
\label{fig:kuangjia}
\end{figure*}

Shah and Koltun~\cite{shah2017robust} jointly solve the tasks of clustering and dimensionality reduction by efficiently optimizing a continuous global objective based on robust statistics, which allows heavily mixed clusters to be untangled. Following this method, a deep continuous clustering approach is suggested in~\cite{shah2018deep}, where the autoencoder parameters and a set of representatives defined against each data-point are simultaneously optimized. The convex clustering approach proposed by~\cite{chi2015splitting} optimizes the representatives by minimizing the distances between each representative and its associated data-point. Non-convex objectives are involved to penalize for the pairwise distances between the representatives.

Furthermore, to improve the performance of clustering, some methods combine convolutional layers with fully connected layers. Joint Unsupervised Learning (JULE)~\cite{yang2016joint} jointly optimizes a convolutional neural network with the clustering parameters in a recurrent manner using an agglomerative clustering approach, where image clustering is conducted in the forward pass and representation learning is performed in the backward pass. Dizaji~\cite{dizaji2017deep} proposes DEPICT, a method that trains a convolutional auto-encoder with a softmax layer stacked on-top of the encoder. The softmax entries represent the assignment of each data-point to one cluster. VaDE~\cite{jiang2016variational} is a variational autoencoder method for deep embedding, and combines a Gaussian Mixture Model for clusering. In \cite{ji2017deep}, a deep autoencoder is trained to minimize a reconstruction loss together with a self-expressive layer. This objective encourages a sparse representation of the original data. Zhou \etal \cite{zhou2018deep} presents a deep adversarial subspace clustering (DASC) method to learn more favorable representations and supervise sample representation learning by adversarial deep learning~\cite{li2018self}. However, the results of reconstruction through low-dimensional representations are often very blurry. One possible way is to train a discriminator with adversarial learning but it can further increase the difficulty of training. Comparatively, our method introduces a relative reconstruction loss and mutual information estimation to obtain more discriminative representations, and jointly optimize the autoencoder network and the deep spectral clustering network for optimal clustering.

\section{Methodology}
As aforementioned, our framework consists of two main components: a dual autoencoder and a deep spectral clustering network. The dual autoencoder, which reconstructs the inputs using the latent representations and their noise versions, is introduced to make the latent representations more robust. In addition, the mutual information estimation between the inputs and the latent representations is applied to preserve the input information as much as possible. Then we utilize the deep spectral clustering network to embed the latent representations into the eigenspace and subsequently clustering is performed. The two networks are merged into a unified framework and jointly optimized with $KL$ divergence. The framework is shown in Fig.~\ref{fig:kuangjia}.

Let $X = \{x_1,...,x_n\}$ denote the input samples, $Z =\{z_1,...,z_n\}$ denote their corresponding latent representations where $z_i=f(x_i;\theta_e)\in \mathbb{R}^d$ is learned by the encoder $\mathbb{E}$. The parameters of the encoder are defined by $\theta_e$, and $d$ is the feature dimension. $\tilde{x}_{z_i}=g(z_i;\theta_d)$ represents the reconstructed data point, which is the output of the decoder $\mathbb{D}$, and the parameters of the decoder are denoted by $\theta_d$. We adopt a deep spectral clustering network $\mathbb{C}$ to map $z_i$ to $y_i=c(z_i;\theta_y)\in \mathbb{R}^K $, where $K$ is the number of clusters.
\subsection{Discriminative latent representation}
We first train the dual autoencoder network to embed the inputs into a latent space. Based on the original reconstruction loss, we add a noise-disturbed reconstruction loss to learn the decoder network. In addition, we introduce the maximization of mutual information~\cite{hjelm2018learning} to the learning procedure of the encoder network, so that the network can obtain more robust representations.

\noindent\textbf{Encoder:} Feature extraction is the major step in clustering and a good feature can effectively improve clustering performance. However, a single reconstruction loss cannot well guarantee the quality of the latent representations. We hope that the representations will help us to identify the sample from the inputs, which means it is the most unique information extracted from the inputs. Mutual information measures the essential correlation between two samples and can effectively estimate the similarity between features $Z$ and inputs $X$. The definition of mutual information is defined as:
\begin{equation}
\begin{split}
I(X,Z)=&\iint p(z|x)p(x)\log\frac{p(z|x)}{p(z)}dxdz\\
      =&KL(p(z|x)p(x)||p(z)p(x)),
\label{eqn: eq1}
\end{split}
\end{equation}
where $p(x)$ is the distribution of the inputs, $p(z|x)$ is the distribution of the latent representations, and the distribution of latent space $p(z)$ can be calculated by $p(z)=\int p(z|x)p(x)dx$. The mutual information is expected to be as large as possible when training the encoder network, hence we have:
\begin{equation}
\begin{split}
p(z|x)=\max_{\theta_e}I(X,Z).
\label{eqn: eq3}
\end{split}
\end{equation}
In addition, the learned latent representations are required to obey the prior distribution of the standard normal distribution with $KL$ divergence. This is beneficial to make the latent space more regular. The distribution difference between $p(z)$ and its prior $q(z)$ is defined as.
\begin{equation}
\begin{split}
KL(p(z)||q(z))=\int p(z)log\frac{p(z)}{q(z)}dz.
\label{eqn: eq4}
\end{split}
\end{equation}
According to Eqs.~\eqref{eqn: eq3} and \eqref{eqn: eq4}, we have:
\begin{equation}
\begin{split}
p(z|x)=\min_{\theta_e}\left\{-\iint p(z|x)p(x)\log\frac{p(z|x)}{p(z)}dxdz\right.\\\left. +\alpha\int p(z)\log\frac{p(z)}{q(z)}dz\right\}.
\label{eqn: eq5}
\end{split}
\end{equation}
It can be further rewritten as:
\begin{equation}
\begin{split}
p(z|x)=\min_{\theta_e}\left\{\iint p(z|x)p(x)[-(\alpha+1)\log\frac{p(z|x)}{p(z)}\right.\\\left. +\alpha\log\frac{p(z|x)}{q(z)}]dxdz\right\}.
\label{eqn: eq2}
\end{split}
\end{equation}
According to Eq.~\eqref{eqn: eq1}, the Eq. \eqref{eqn: eq2} can be viewed as:
\begin{equation}
\begin{split}
p(z|x)=\min_{\theta_e}&\left\{-\beta I(X,Z)\right.\\&\left.+\gamma\mathbb{E}_{x\thicksim p(x)}[KL(p(z|x)||q(z))]\right\}.
\label{eqn: eq6}
\end{split}
\end{equation}
Unfortunately, $KL$ divergence is unbounded. Instead of using $KL$ divergence, $JS$ divergence is adopted for mutual information maximization:
\begin{equation}
\begin{split}
p(z|x)=\min_{\theta_e}&\left\{-\beta JS(p(z|x)p(x),p(z)p(x))\right.\\&\left.+\gamma\mathbb{E}_{x\thicksim p(x)}[KL(p(z|x)||q(z))]\right\}.
\label{eqn: eq7}
\end{split}
\end{equation}

We have known that the variational estimation of $JS$ divergence~\cite{nowozin2016f} is defined as:
\begin{equation}
\begin{split}
JS(p(x)||q(x))=\max_T&(\mathbb{E}_{x\thicksim p(x)}[\log\sigma(T(x))]\\+&\mathbb{E}_{x\thicksim q(x)}[\log(1-\sigma(T(x)))]).
\label{eqn: eq71}
\end{split}
\end{equation}
where $T(x)=\log\frac{2p(x)}{p(x)+q(x)}$~\cite{nowozin2016f}. Here $p(z|x)p(x)$ and $p(z)p(x)$ are utilized to replace $p(x)$ and $q(x)$. As a result, Eq. \eqref{eqn: eq7} can be defined as:
\begin{equation}
\begin{split}
p(z|x)=\min_{\theta_e}&\left\{-\beta(\mathbb{E}_{(x,z)\thicksim p(z|x)p(x)}[\log\sigma(T(x,z))]\right.\\
+&\mathbb{E}_{(x,z)\thicksim p(z)p(x)}[\log(1-\sigma(T(x,z)))])\\
+&\left.\gamma\mathbb{E}_{x\thicksim p(x)}[KL(p(z|x)||q(z))]\right\}.
\label{eqn: eq8}
\end{split}
\end{equation}

Negative sampling estimation~\cite{hjelm2018learning}, which is the process of using a discriminator to distinguish the real and noisy samples to estimate the distribution of real samples, is generally utilized to solve the problem in Eq.~\eqref{eqn: eq8}. $\sigma(T(x,z))$ is a discriminator, where $x$ and its latent representation $z$ together form a positive sample pair. We randomly select $z_t$ from the disturbed batch to construct a negative sample pair according to $x$. Note that Eq.~\eqref{eqn: eq8} represents the global mutual information between $X$ and $Z$.

\begin{figure}[t]
\centering
\includegraphics[height=5.1cm,width=0.45\textwidth]{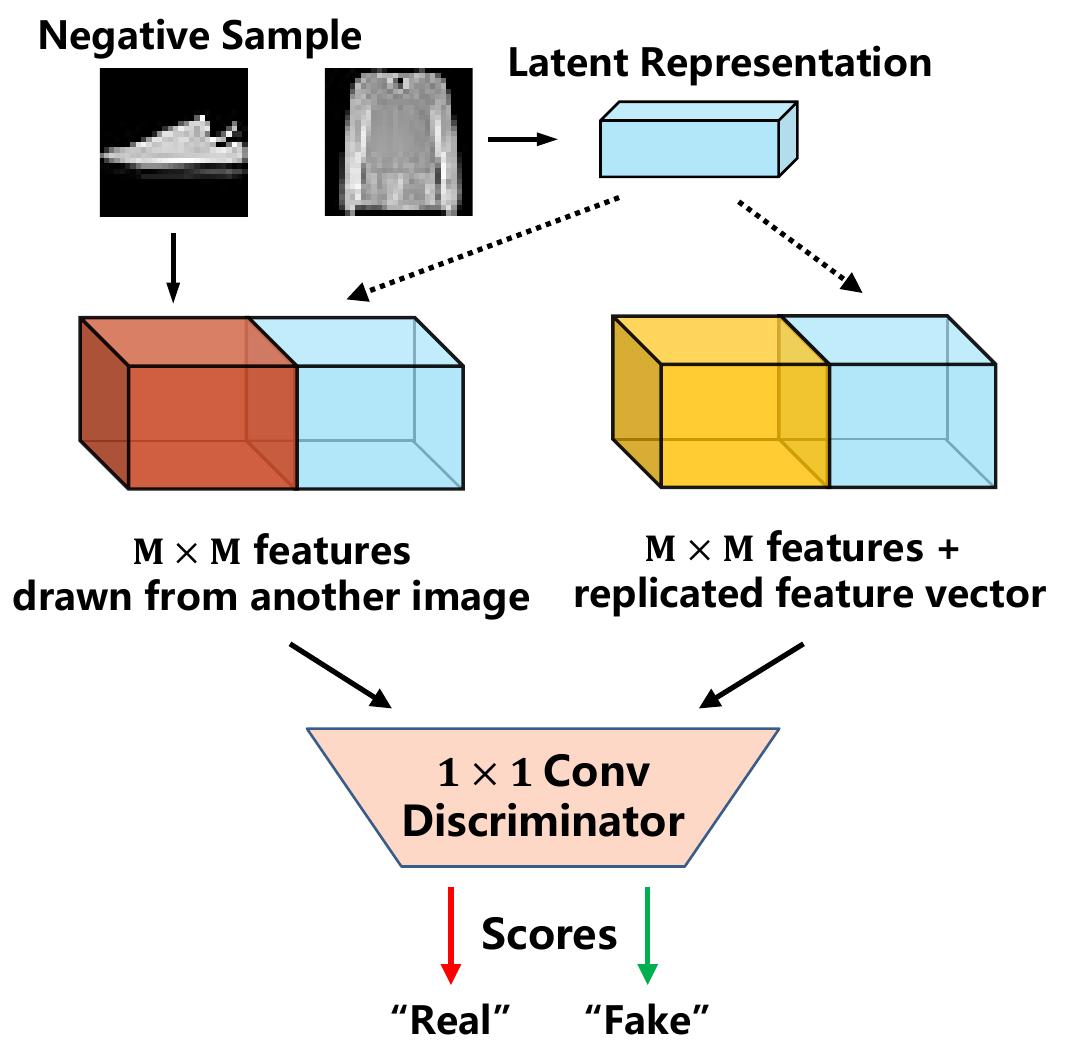}\\
\caption{Local mutual information estimation.}
\label{fig:caiyang}
\end{figure}

Furthermore, we extract the feature map from the middle layer of the convolutional network, and construct the relationship between the feature map and the latent representation, which is the local mutual information. The estimation method plays the same role as global mutual information. The middle layer feature are combined with the latent representation to obtain a new feature map. Then a $1\times1$ convolution is considered as the estimation network of local mutual information, as shown in Fig.~\ref{fig:caiyang}. The selection method of negative samples is the same as global mutual information estimation. Therefore, the objective function that needs to be optimized can be defined as:
\begin{equation}
\begin{split}
L_e=&-\beta(\mathbb{E}_{(x,z)\thicksim p(z|x)p(x)}[\log\sigma(T_1(x,z))]\\
&+\mathbb{E}_{(x,z)\thicksim p(z)p(x)}[\log(1-\sigma(T_1(x,z)))])\\
&-\frac{\beta}{hw}\Sigma_{i,j}(\mathbb{E}_{(x,z)\thicksim p(z|x)p(x)}[\log\sigma(T_2(C_{ij},z))]\\
&+\mathbb{E}_{(x,z)\thicksim p(z)p(x)}[\log(1-\sigma(T_2(C_{ij},z)))])\\
&+\gamma\mathbb{E}_{x\thicksim p(x)}[KL(p(z|x)||q(z))],
\label{eqn: eq9}
\end{split}
\end{equation}
where $h$ and $w$ represent the height and width of the feature map. $C_{ij}$ represents the feature vector of the middle feature map at coordinates $(i,j)$ and $q(z)$ is the standard normal distribution.

\noindent\textbf{Decoder:} In the existing decoder networks, the reconstruction loss is generally a suboptimal scheme for clustering, due to the natural trade-off between the reconstruction and the clustering tasks. The reconstruction loss mainly depends on the two parts: the distribution of the latent representations and the generative capacity of decoder network. However, the generative capacity of the decoder network is not required in the clustering task. Our real goal is not to obtain the best reconstruction results, but to get more discriminative features for clustering. We directly use noise disturbance in the latent space to discard known nuisance factors from the latent representations. Models trained in this fashion become robust by exclusion rather than inclusion, and are expected to perform well on clustering tasks, where even the inputs contain unseen nuisance~\cite{jaiswal2018unsupervised}. A noisy-transformer $\psi$ is utilized to convert the latent representations $Z$ into their noisy versions $\hat{Z}$, and then the decoder reconstructs the inputs from $\hat{Z}$ and $Z$. The reconstruction results can be defined as $\tilde{x}_{\hat{z}_i}=g(\hat{z}_i;\theta_d)$ and $\tilde{x}_{z_i}=g(z_i;\theta_d)$, and the relative reconstruction loss can be written as:
\begin{equation}
\begin{split}
L_r(\tilde{x}_{\hat{z}_i},\tilde{x}_{z_i})=\parallel\tilde{x}_{\hat{z}_i}-\tilde{x}_{z_i}\parallel_F^2,
\label{eqn: eq10}
\end{split}
\end{equation}
where $\parallel\cdot\parallel_F$ stands for the Frobenius norm.
We also use the original reconstruction loss to ensure the performance of the decoder network and consider $\psi$ as multiplicative Gaussian noise. The complete reconstruction loss can be defined as:
\begin{equation}
\begin{split}
L_r =\parallel\tilde{x}_{\hat{z}_i}-\tilde{x}_{z_i}\parallel_F^2+\delta \parallel x-\tilde{x}_{z_i}\parallel_F^2.
\label{eqn: eq11}
\end{split}
\end{equation}
where $\delta$ stands for the strength of different reconstruction loss.

Hence, by considering all the items, the total loss of the autoencoder network can be defined as:
\begin{equation}
\begin{split}
\min_{\theta_d,\theta_e}  L_r+ L_e.
\label{eqn: eq20}
\end{split}
\end{equation}
\subsection{Deep Spectral Clustering}
The learned autoencoder parameters $\theta_e$ and $\theta_d$ are considered as an initial condition in the clustering phase. Spectral clustering can effectively use the relationship between samples to reduce intra-class differences, and produce better clustering results than $K$-means. In this step, we first adopt the autoencoder network to learn the latent representations. Next, a spectral clustering method is used to embed the latent representations into the eigenspace of their associated graph Laplacian matrix~\cite{liu2010large}. All the samples will be subsequently clustered in this space. Finally, both the autoencoder parameters and clustering objective are jointly optimized.

Specifically, we first utilize the latent representations $Z$ to construct the non-negative affinity matrix $W$:
\begin{equation}
\begin{split}
W_{i,j}=e^{-\frac{\parallel z_i-z_j\parallel^2}{2\sigma^2}}.
\label{eqn: eq12}
\end{split}
\end{equation}
The loss function of spectral clustering is defined as:
\begin{equation}
\begin{split}
L_c=\mathbb{E}[W_{i,j}\parallel y_i-y_j\parallel^2],
\label{eqn: eq13}
\end{split}
\end{equation}
where $y_i$ is the output of the network. When we adopt the general neural network to output $y$, we randomly select a minibatch of $m$ samples at each iteration and thus the loss function can be defined as:
\begin{equation}
\begin{split}
L_c=\frac{1}{m^2}\sum_{i,j=1}^mW_{i,j}\parallel y_i-y_j\parallel^2.
\label{eqn: eq15}
\end{split}
\end{equation}

In order to prevent that all points are grouped into the same cluster in network maps, the output $y$ is required to be orthonormal in expectation. That is to say:
\begin{equation}
\begin{split}
\frac{1}{m}Y^TY=I_{k\times k},
\label{eqn: eq17}
\end{split}
\end{equation}
where $Y$ is a $m\times k$ matrix of the outputs whose $i$th row is $y_i^T$. The last layer of the network is utilized to enforce the orthogonality~\cite{shaham2018spectralnet} constraint. This layer gets input from $K$ units, and acts as a linear layer with $K$ outputs, in which the weights are required to be orthogonal, producing the orthogonalized output $Y$ for a minibatch. Let $\tilde{Y}$ denote the $m\times k$ matrix containing the inputs to this layer for $Z$, a linear map that orthogonalizes the columns of $\tilde{Y}$ is computed through its QR decomposition. Since integrated $A^\top A$ is full rank for any matrix $A$, the QR decomposition can be obtained by the Cholesky decomposition:
\begin{equation}
\begin{split}
A^\top A=BB^\top,
\label{eqn: eq18}
\end{split}
\end{equation}
where $B$ is a lower triangular matrix, and $Q=A(B^{-1})^\top$. Therefore, in order to orthogonalize $\tilde{Y}$, the last layer multiplies $\tilde{Y}$ from the right by $\sqrt{m}(L^{-1})^T$. Actually, $\tilde{L}$ can be obtained from the Cholesky decomposition of $\tilde{Y}$ and the $\sqrt{m}$ factor is needed to satisfy Eq. \eqref{eqn: eq17}.

We unify the latent representation learning and the spectral clustering using $KL$ divergence. In the clustering phase, the last term of Eq.~\eqref{eqn: eq9} can be rewritten as:
\begin{equation}
\begin{split}
\mathbb{E}_{x\thicksim p(x)}[KL(p((y,z)|x)||q(y,z))],
\label{eqn: eq19}
\end{split}
\end{equation}
where $p((y,z)|x)=p(y|z)p(z|x)$ and $q(y,z)=q(z|y)q(y)$. Note $q(z|y)$ is a normal distribution with mean $\mu_y$ and variance $1$. Therefore, the overall loss of the autoencoder and the spectral clustering network is defined as:
\begin{equation}
\begin{split}
\min_{\theta_d,\theta_e,\theta_c}  L_r+ L_e+ L_c.
\label{eqn: eq20}
\end{split}
\end{equation}

Finally, we jointly optimize the two networks until convergence to obtain the desired clustering results.
\section{Experiments}
In this section, we evaluate the effectiveness of the proposed clustering method in five benchmark datasets, and then compare the performance with several state-of-the-arts.
\subsection{Datasets}
\begin{figure}[t]
 \centering
  \subfigure[MNIST]{
  \centering
  \label{fig:subfigs:YaleB} 
  \includegraphics[height=2.1cm,width=7.5cm]{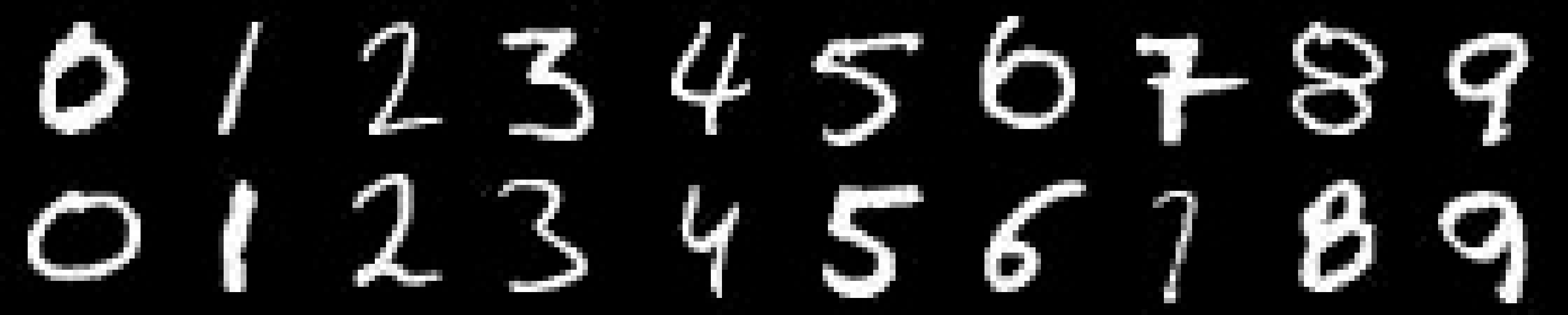}}
  \hspace{.123in}
  \subfigure[Fashion-Mnist]{
  \centering
  \label{fig:subfigs:COIL20} 
  \includegraphics[height=2.1cm,width=7.5cm]{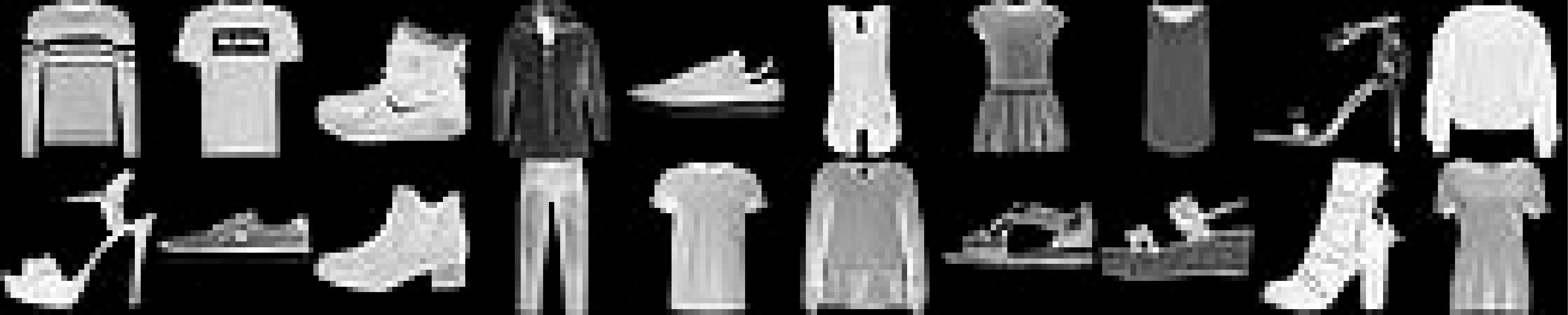}}
   \caption{The image samples from the benchmark datasets used in our experiments}
   \label{fig: benchmark datasets} 
 \end{figure}

In order to show that our method works well with various kinds of datasets, we choose the following image datasets. Considering that clustering tasks are fully unsupervised, we concatenate the
training and testing samples when applicable. MNIST-\emph{full}~\cite{lecun1998gradient}: A dataset containing a total of 70,000 handwritten digits with 60,000 training and 10,000 testing samples, each being a 32$\times$32 monochrome image. MNIST-\emph{test}: A dataset only consists of the testing part of MNIST-\emph{full} data. USPS: A handwritten digits dataset from the USPS postal service, containing 9,298 samples of 16$\times$16 images. Fashion-MNIST~\cite{xiao2017fashion}: This dataset has the same number of images and the same image size with MNIST, but it is fairly more complicated. Instead of digits, it consists of various types of fashion products. YTF: We adopt the first 41 subjects of YTF dataset and the images are first cropped and resized to $55\times55$. Some image samples are shown in Fig. \ref{fig: benchmark datasets}. The brief descriptions of the datasets are given in Tab. \ref{table:tab1}.

\begin{table}[t]
\centering
 \caption{Description of Datasets}
 \begin{tabular}{|c|c|c|c|}
   \hline
  Dataset& Samples & Classes& Dimensions\\
  \hline
 MNIST-\emph{full} & 70,000 & 10 & 1$\times$28$\times$28\\\hline
 MNIST-\emph{test} & 10,000 & 10 & 1$\times$28$\times$28\\\hline
 USPS & 9298 & 10 & 1$\times$16$\times$16\\\hline
 Fashion-Mnist & 70,000 & 10 & 1$\times$28$\times$28\\\hline
 YTF & 10,000 & 41 & 3$\times$55$\times$55\\\hline
 \end{tabular}
 \label{table:tab1}
\end{table}
\subsection{Clustering Metrics}
To evaluate the clustering results, we adopt two standard evaluation metrics: Accuracy (ACC) and Normalized Mutual Information (NMI) \cite{xu2003document}.

The best mapping between cluster assignments and true labels is computed using the Hungarian algorithm to measure accuracy~\cite{kuhn1955hungarian}. For completeness, we define ACC by:
\begin{equation}
\begin{split}
ACC = \max_m\frac{\sum_{i=1}^n\mathbf{1}\{l_i=m(c_i)\}}{n},
\label{eqn: eq21}
\end{split}
\end{equation}
where $l_i$ and $c_i$ are the true label and predicted cluster of data point $x_i$.

NMI calculates the normalized measure of similarity between two labels of the same data, which is defined as:
\begin{equation}
\begin{split}
NMI=\frac{I(l;c)}{max\{H(l),H(c)\}},
\label{eqn: eq22}
\end{split}
\end{equation}
where $I(l,c)$ denotes the mutual information between true label $l$ and predicted cluster $c$, and $H$ represents their entropy. Results of NMI do not change by permutations of clusters (classes), and they are normalized to the range of [0, 1], with 0 meaning no correlation and 1 exhibiting perfect correlation.
\begin{table*}[!ht]
  \centering
  \caption{Description the structure of the autoencoder network}
    \begin{tabular}{c|cccccc}
    \toprule
    Method& encoder-1/decoder-4&encoder-2/decoder-3&encoder-3/decoder-2&encoder-4/decoder-1 \\\hline
    MNIST&3$\times$3$\times$16&3$\times$3$\times$16&3$\times$3$\times$32&3$\times$3$\times$32\\
    USPS&3$\times$3$\times$16&3$\times$3$\times$32&-&-\\
    Fashion-Mnist&3$\times$3$\times$16&3$\times$3$\times$16&3$\times$3$\times$32&3$\times$3$\times$32\\
    YTF&5$\times$5$\times$16&5$\times$5$\times$16&5$\times$5$\times$32&5$\times$5$\times$32\\
    \bottomrule
    \end{tabular}
    \label{tab:network}
\end{table*}
\subsection{Implementation Details}
In our experiments, we set $\beta=0.01$, $\gamma=1$, and $\delta=0.5$. The channel numbers and kernel sizes of the autoencoder network are shown in Tab.~\ref{tab:network}, and the dimension of latent space is set to 120. The deep spectral clustering network consists of four fully connected layers, and we adopt ReLU~\cite{maas2013rectifier} as the non-linear activations. We construct the original weight matrix $W$ with probabilistic $K$-nearest neighbors for each dataset. The weight $W_{ij}$ is calculated as nearest-neighbor graph \cite{gu2009co}, and the number of neighbors is set to 3.
\subsection{Comparison Methods}
We compare our clustering model with several baselines, including $K$-means~\cite{macqueen1967some}, spectral clustering with normalized cuts (SC-Ncut)~\cite{shi2000normalized}, large-scale spectral clustering (SC-LS)~\cite{chen2011large}, NMF~\cite{cai2009locality}, graph degree linkage-based agglomerative clustering (AC-GDL)~\cite{zhang2013agglomerative}. In addition, we also evaluate the performance of our method with several state-of-the-art clustering algorithms based on deep learning, including deep adversarial subspace clustering (DASC)~\cite{zhou2018deep}, deep embedded clustering (DEC)~\cite{xie2016unsupervised}, variational deep embedding (VaDE)~\cite{jiang2016variational}, joint unsupervised learning (JULE)~\cite{yang2016joint}, deep embedded regularized clustering (DEPICT)~\cite{dizaji2017deep}, improved deep embedded clustering with locality preservation (IDEC)~\cite{guo2017improved}, deep spectral clustering with a set of nearest neighbor pairs (SpectralNet)~\cite{shaham2018spectralnet}, clustering with GAN (ClusterGAN)~\cite{mukherjee2018clustergan} and GAN with the mutual information (InfoGAN)~\cite{chen2016infogan}.
\begin{table*}[!ht]
  \centering
  \caption{Clustering performance of different algorithms on five datasets based on ACC and NMI}
  \label{tab:performance_comparison}
    \begin{tabular}{|c|c|c|c|c|c|c|c|c|c|c|}
    \hline
    \multirow{2}{*}{Method}& \multicolumn{2}{|c|}{MNIST-\emph{full}}&\multicolumn{2}{|c|}{MNIST-\emph{test}}&\multicolumn{2}{|c|}{USPS}&\multicolumn{2}{|c|}{Fashion-10}&\multicolumn{2}{|c|}{YTF} \\ \cline{2-11}
    &NMI&ACC&NMI&ACC&NMI&ACC&NMI&ACC&NMI&ACC \\\hline
    K-means~\cite{macqueen1967some}&0.500&0.532&0.501&0.546&0.601&0.668&0.512&0.474&0.776&0.601\\\hline
    SC-Ncut~\cite{shi2000normalized}&0.731&0.656&0.704&0.660&0.794&0.649&0.575&0.508&0.701&0.510\\\hline
    SC-LS~\cite{chen2011large}&0.706&0.714&0.756&0.740&0.755&0.746&0.497&0.496&0.759&0.544\\\hline
    NMF~\cite{cai2009locality}&0.452&0.471&0.467&0.479&0.693&0.652&0.425&0.434&-&-\\\hline
    AC-GDL~\cite{zhang2013agglomerative}&0.017&0.113&0.864&0.933&0.825&0.725&0.010&0.112&0.622&0.430\\\hline\hline
    DASC~\cite{zhou2018deep}&0.784$^*$&0.801$^*$&0.780&0.804&-&-&-&-&-&-\\\hline
    DEC~\cite{xie2016unsupervised}&0.834$^*$&0.863$^*$&0.830$^*$&0.856$^*$&0.767$^*$&0.762$^*$&0.546$^*$&0.518$^*$&0.446$^*$&0.371$^*$\\\hline
    VaDE~\cite{jiang2016variational}&0.876&0.945&-&-&0.512&0.566&0.630&0.578&-&-\\\hline
    JULE~\cite{yang2016joint}&0.913$^*$&0.964$^*$&0.915$^*$&0.961$^*$&\textbf{0.913}&\textbf{0.950}&0.608&0.563&0.848&0.684\\\hline
    DEPICT~\cite{dizaji2017deep}&0.917$^*$&0.965$^*$&0.915$^*$&0.963$^*$&0.906&0.899&0.392&0.392&0.802&0.621\\\hline
    IDEC~\cite{guo2017improved}&0.867$^*$&0.881$^*$&0.802&0.846&0.785$^*$&0.761$^*$&0.557&0.529&-&-\\\hline
    SpectralNet~\cite{shaham2018spectralnet}&0.814&0.800&0.821&0.817&-&-&-&-&0.798&0.685\\\hline
    InfoGAN~\cite{chen2016infogan} &0.840&0.870&-&-&-&-&0.590&0.610&-&-\\\hline
    ClusterGAN~\cite{mukherjee2018clustergan}&0.890&0.950&-&-&-&-&0.640&0.630&-&-\\\hline
    Our Method&\textbf{0.941}&\textbf{0.978}&\textbf{0.946}&\textbf{0.980}&0.857&0.869&\textbf{0.645}&\textbf{0.662}&\textbf{0.857}&\textbf{0.691}\\\hline
    \end{tabular}
\end{table*}
\subsection{Evaluation of Clustering Algorithm}
We run our method with 10 random trials and report the average performance, the error range is no more than 2\%. In terms of the compared methods, if the results of their methods on some datasets are not reported, we run the released code with hyper-parameters mentioned in their papers, and the results are marked by (*) on top. When the code is not publicly available, or running the released code is not practical, we put dash marks (-) instead of the corresponding results.

The clustering results are shown in Tab.~\ref{tab:performance_comparison}, where the first five are conventional clustering methods. In the table, we can notice that our proposed method outperforms the competing methods on these benchmark datasets. We observe that the proposed method can improve the clustering performance whether in digital datasets or in other product dataset. Especially when performing on the object dataset MNIST-\emph{test}, the clustering accuracy is over 98\%. Specifically, it exceeds the second best DEPICT which is trained on the noisy versions of the inputs by 1.6\% and 3.1\% on ACC and NMI respectively. Moreover, our method achieves much better clustering results than several classical shallow baselines. This is because compared with shallow methods, our method uses a multi-layer convolutional autoencoder as the feature extractor and adopts deep clustering network to obtain the most optimal clustering results. The Fashion-MNIST dataset is very difficult to deal with due to the complexity of samples, but our method still harvests good results.

\begin{figure}[!ht]
 \centering
  \subfigure[]{
  \centering
  \label{fig:subfigs:c2} 
  \includegraphics[height=2.8cm,width=4cm,]{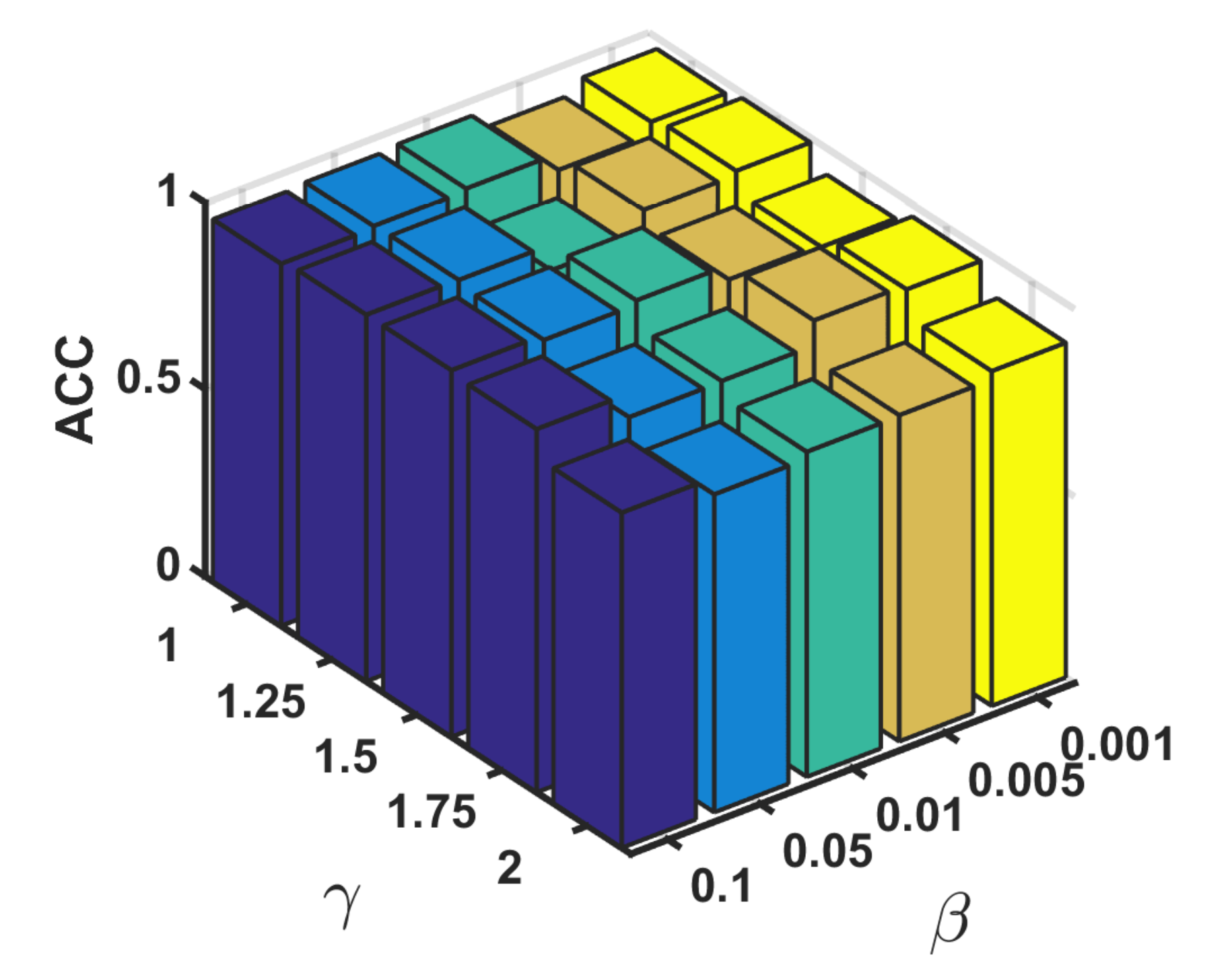}}
   \subfigure[]{
  \centering
  \label{fig:subfigs:s2} 
  \includegraphics[height=2.8cm,width=4cm]{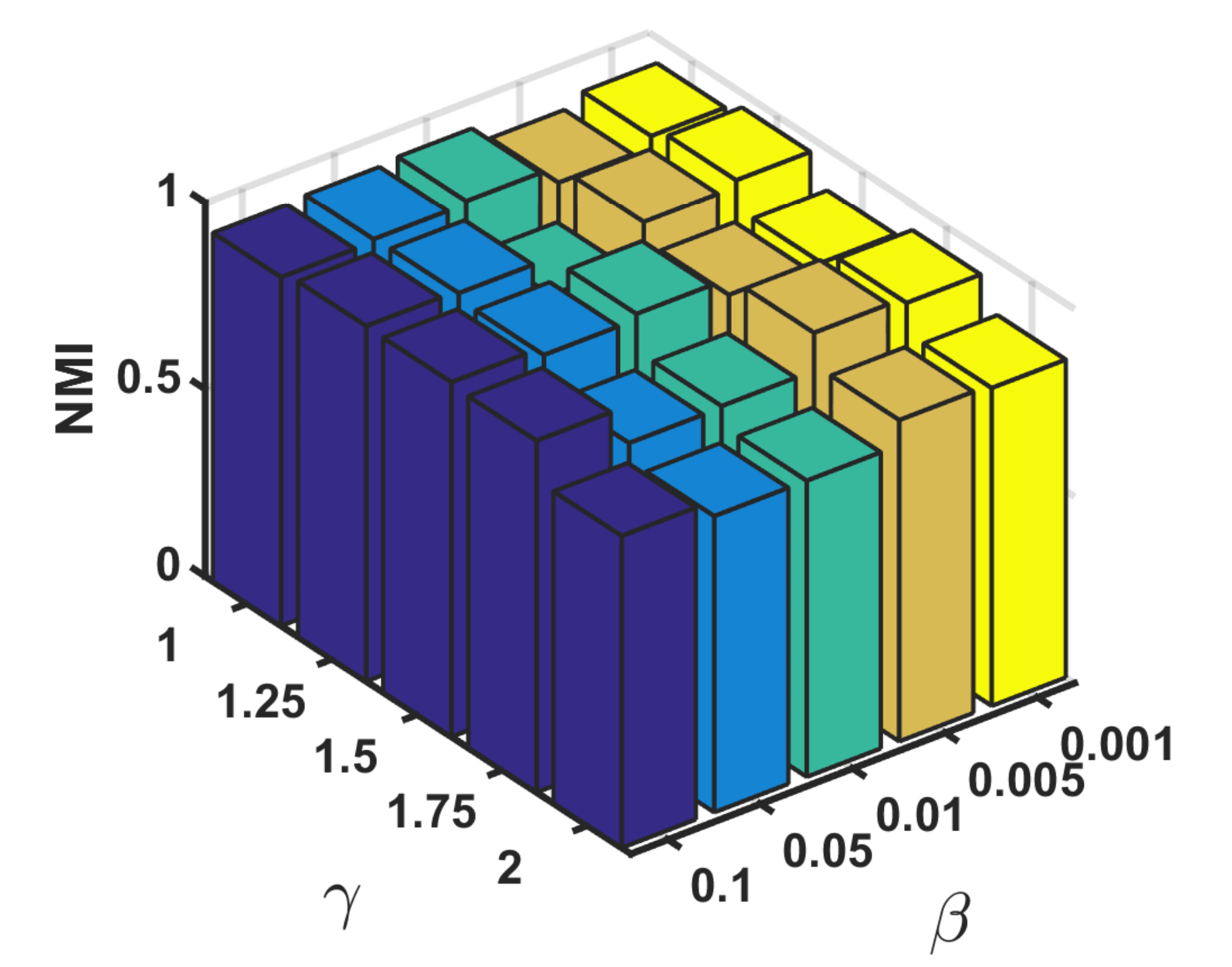}}
   \caption{ACC and NMI of Our method with different $\beta$ and $\gamma$ on MNIST dataset}
   \label{fig: fenlei} 
 \end{figure}
We also investigate the parameter sensitivity on MNIST-\emph{test}, and the results are shown in Fig.~\ref{fig: fenlei}, where Fig.~\ref{fig:subfigs:c2} represents the results of ACC from different parameters and Fig.~\ref{fig:subfigs:s2} is the results of NMI. It intuitively demonstrates that our method maintains acceptable results with most parameter combinations and has relative stability.
\subsection{Evaluation of Learning Approach}
 \begin{table*}[!ht]
  \centering
  \caption{Clustering performance with different strategies on five datasets based on ACC and NMI}
  \label{tab:multi}
    \begin{tabular}{|c|c|c|c|c|c|c|c|c|c|c|}
    \hline
    \multirow{2}{*}{Method}& \multicolumn{2}{|c|}{MNIST-\emph{full}}&\multicolumn{2}{|c|}{MNIST-\emph{test}}&\multicolumn{2}{|c|}{USPS}&\multicolumn{2}{|c|}{Fashion-10} &\multicolumn{2}{|c|}{YTF}\\ \cline{2-11}
    &NMI&ACC&NMI&ACC&NMI&ACC&NMI&ACC&NMI&ACC \\\hline
    ConvAE&0.745&0.776&0.751&0.781&0.652&0.698&0.556&0.546&0.642&0.476\\\hline
    ConvAE+MI&0.800&0.835&0.796&0.844&0.744&0.785&0.609&0.592&0.738&0.571\\\hline
    ConvAE+RS&0.803&0.841&0.801&0.850&0.752&0.798&0.597&0.614&0.721&0.558\\\hline
    ConvAE+MI+RS&0.910&0.957&0.914&0.961&0.827&0.831&0.640&0.656&0.801&0.606\\\hline
    ConvAE+MI+RS+SN&0.941&0.978&0.946&0.980&0.857&0.869&0.645&0.662&0.857&0.691\\\hline
    \end{tabular}
    \vspace{-0.4cm}
\end{table*}

\begin{figure*}[!ht]
 \centering
  \subfigure[Raw data]{
  \centering
  \label{fig:subfigs:1} 
  \includegraphics[height=3.0cm,width=3.5cm]{Figure/1.png}}
  \hspace{.01in}
   \subfigure[ConvAE]{
  \centering
  \label{fig:subfigs:2} 
  \includegraphics[height=3.0cm,width=3.5cm]{Figure/2.png}}
  \hspace{.01in}
   \subfigure[DEC]{
  \centering
  \label{fig:subfigs:3} 
  \includegraphics[height=3.0cm,width=3.5cm]{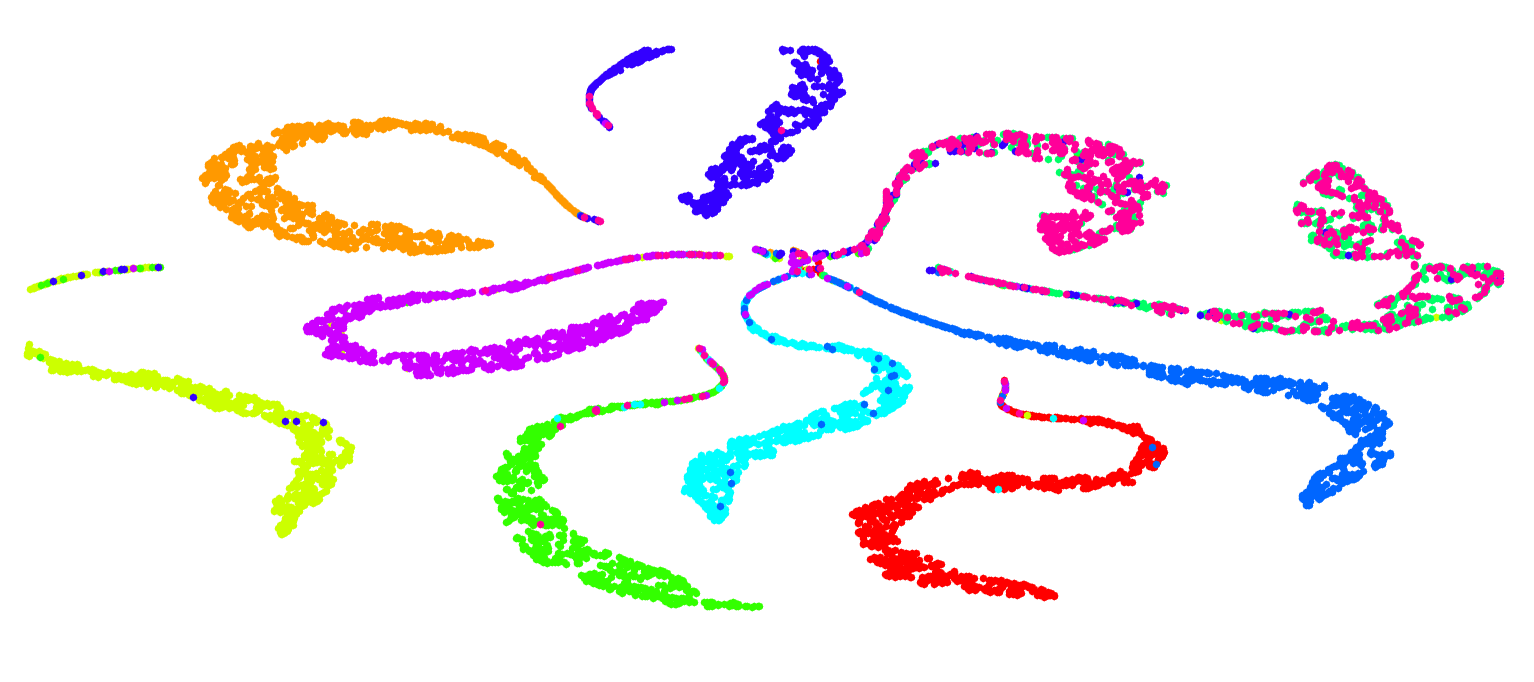}}
    \hspace{.01in}
   \subfigure[SpectralNet]{
  \centering
  \label{fig:subfigs:4} 
  \includegraphics[height=3.0cm,width=3.5cm]{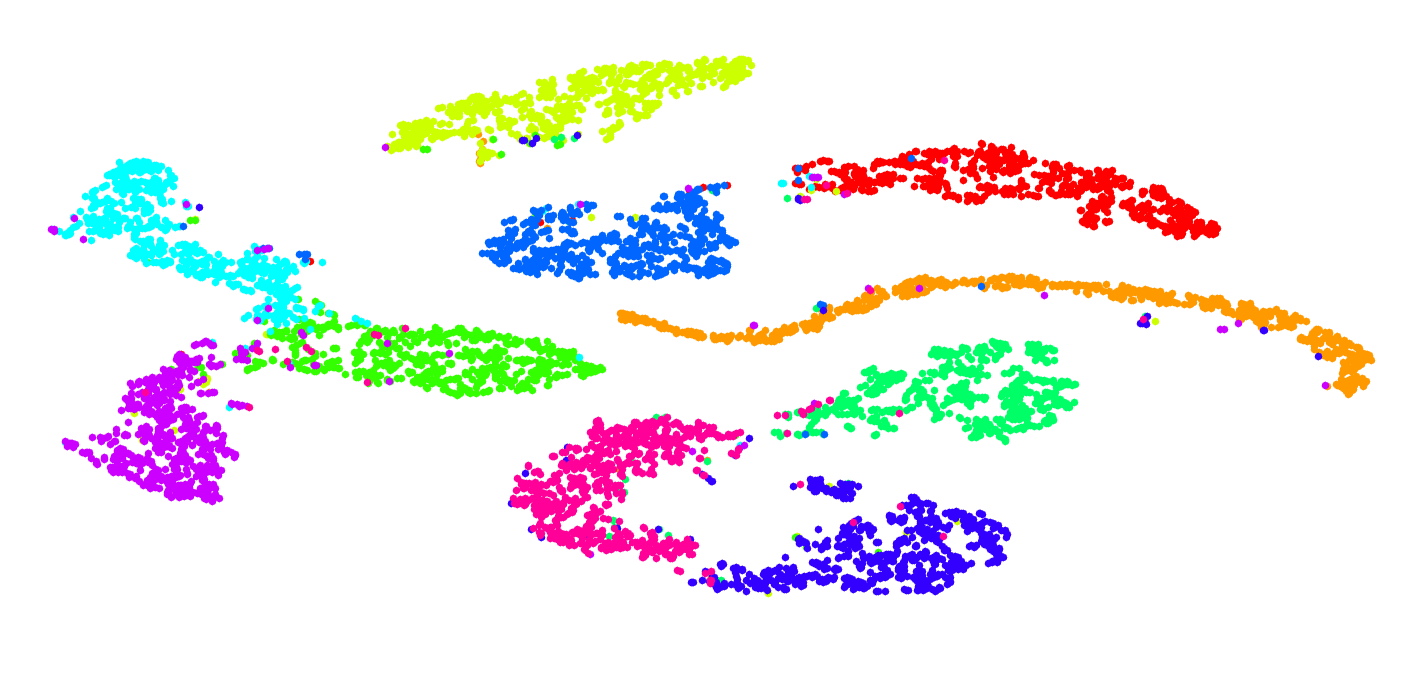}}
    \subfigure[ConvAE+RS]{
  \centering
  \label{fig:subfigs:5} 
  \includegraphics[height=3.0cm,width=3.5cm,]{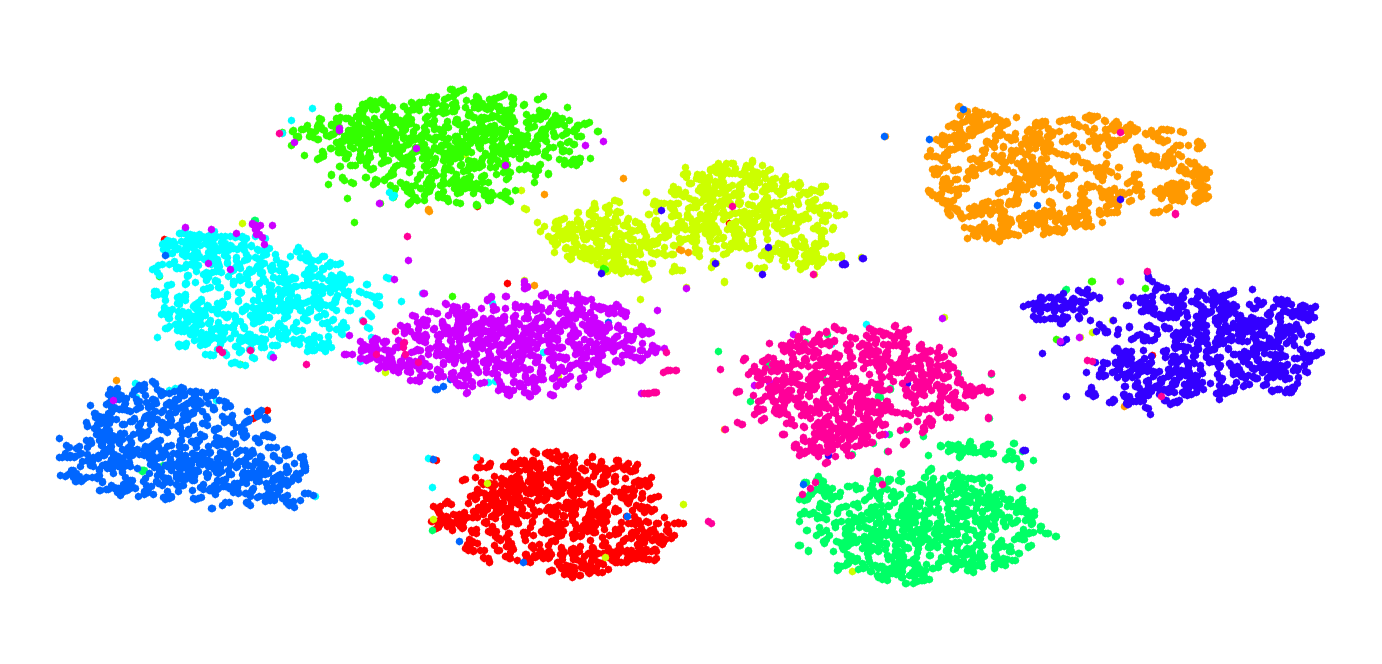}}
  \hspace{.01in}
   \subfigure[ConvAE+MI]{
  \centering
  \label{fig:subfigs:6} 
  \includegraphics[height=3.0cm,width=3.5cm]{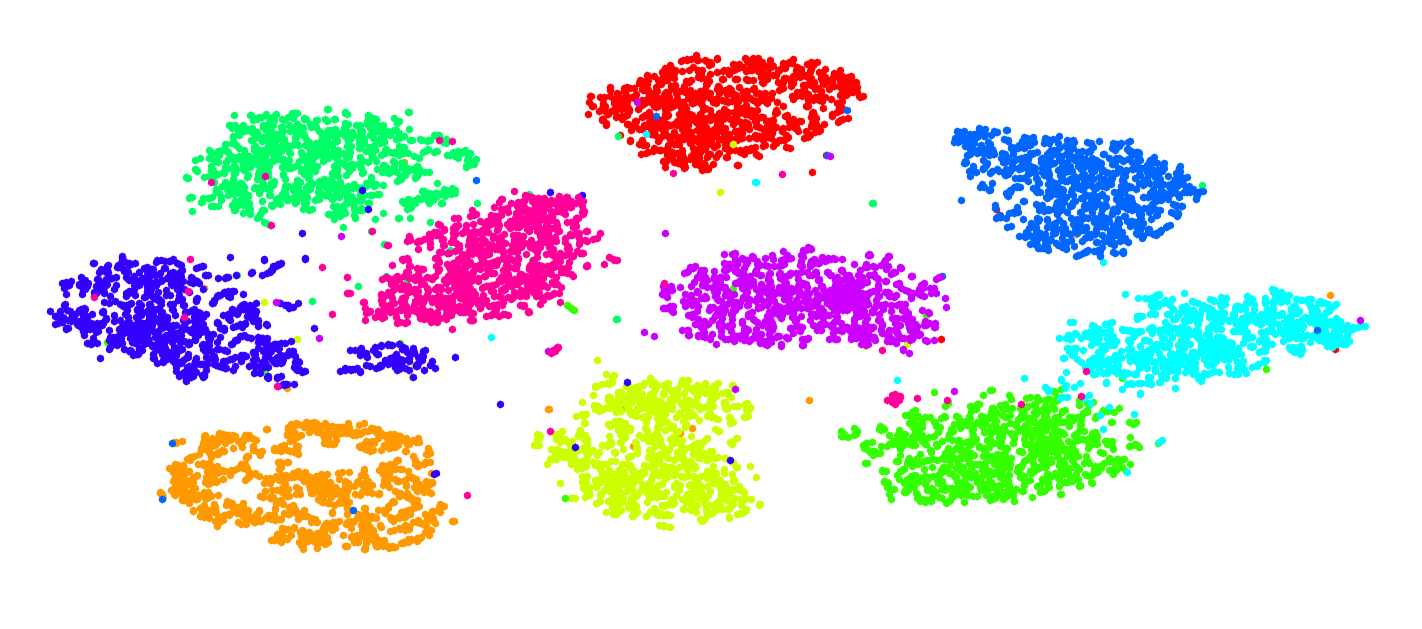}}
  \hspace{.01in}
   \subfigure[ConvAE+RS+MI]{
  \centering
  \label{fig:subfigs:7} 
  \includegraphics[height=3.0cm,width=3.5cm]{Figure/3.png}}
    \hspace{.01in}
   \subfigure[ConvAE+MI+RS+SN]{
  \centering
  \label{fig:subfigs:8} 
  \includegraphics[height=3.0cm,width=3.5cm]{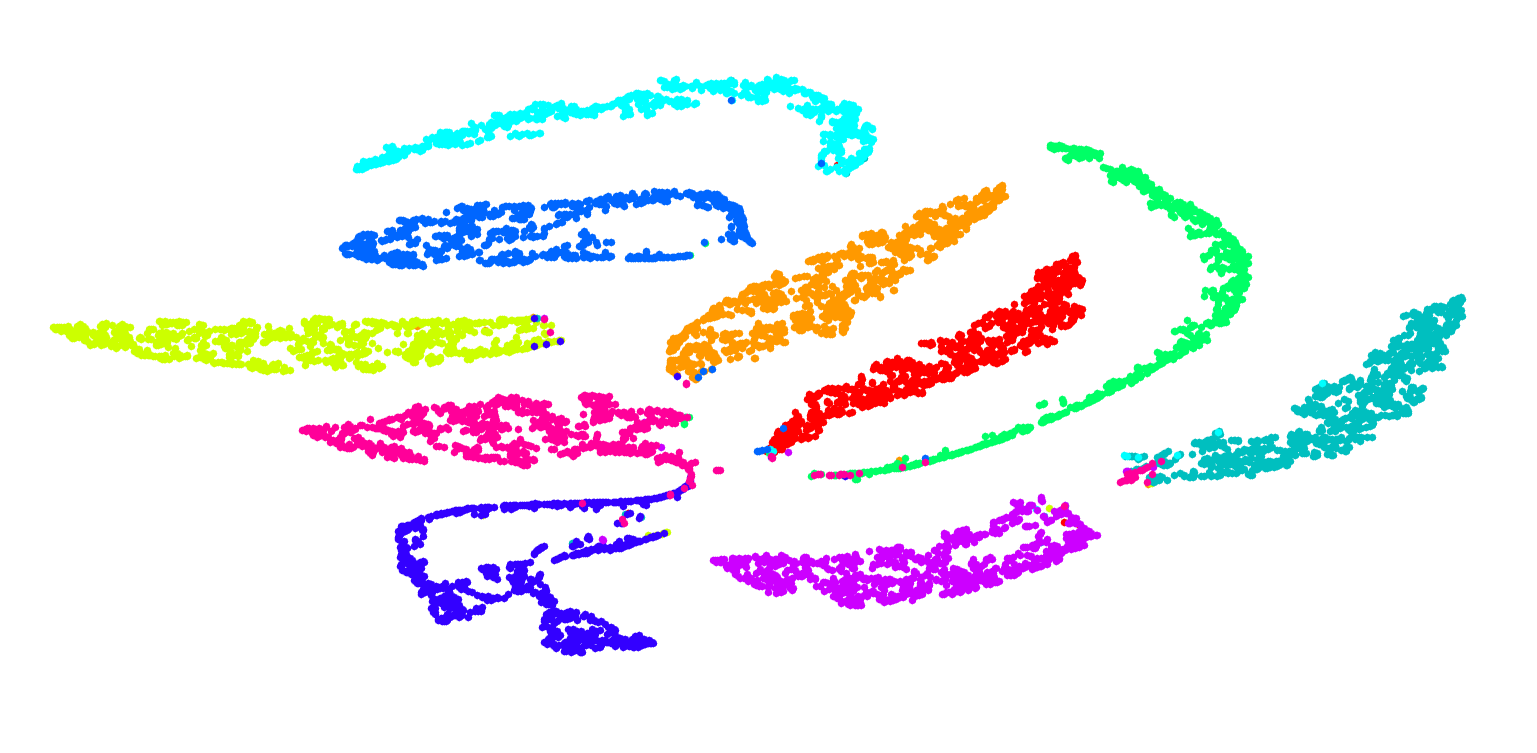}}
   \caption{Visualization to show the discriminative capability of embedding subspaces using MNIST-\emph{test} data.}
   \label{fig: multi} 
 \end{figure*}
We compare different strategies for training our model. For training a multi-layer convolutional autoencoder, we analyze the following four approaches: (1) convolutional autoencoder with original reconstruction loss (ConvAE), (2) convolutional autoencoder with original reconstruction loss and mutual information (ConvAE+MI), (3) convolutional autoencoder with improved reconstruction loss (ConvAE+RS) and (4) convolutional autoencoder with improved reconstruction loss and mutual information (ConvAE+MI+RS). The last one is the joint training of convolutional autoencoder and deep spectral clustering. Tab.~\ref{tab:multi} represents the performance of different strategies for training our model. It clearly demonstrates that each kind of strategy  of our method can improve the accuracy of clustering effectively, especially after adding mutual information and the improved reconstruction loss in the convolutional autoencoder network. Fig.~\ref{fig: multi} demonstrates the importance of our proposed strategy by comparing different data representations of MNIST-\emph{test} data points using $t$-SNE visualization~\cite{maaten2008visualizing}, Fig.~\ref{fig:subfigs:1} represents the space of raw data, Fig.~\ref{fig:subfigs:2} is the data points in the latent subspace of convolution autoencoder,  Fig.~\ref{fig:subfigs:3} and \ref{fig:subfigs:4} are the results of DEC and SpectralNet respectively, and the rest are our proposed model with different strategies. The results demonstrate the latent representations obtained by our method have more clear distribution structure.
\section{Conclusion}
In this paper, we propose an unsupervised deep clustering method with a dual autoencoder network and a deep spectral network. First, the dual autoencoder, which reconstructs the inputs using the latent representations and their noise-contaminated versions, is utilized to establish the relationships between the inputs and the latent representations in order to obtain more robust latent representations. Furthermore, we maximize the mutual information between the inputs and the latent representations, which can preserve the information of the inputs as much as possible. Hence, the features of the latent space obtained by our autoencoder are robust to noise and more discriminative. Finally, the spectral network is fused to a unified framework to cluster the features of the latent space, so that the relationship between the samples can be effectively utilized. We evaluate our method on several benchmarks and experimental results show that our method outperforms those state-of-the-art approaches.
\section{Acknowledgement}
Our work was also supported by the National Natural Science Foundation of China under Grant 61572388, 61703327 and 61602176, the Key R\&D Program-The Key Industry Innovation Chain of Shaanxi under Grant 2017ZDCXL-GY-05-04-02, 2017ZDCXL-GY-05-02 and 2018ZDXM-GY-176, and the National Key R\&D Program of China under Grant 2017YFE0104100.

{\small
\bibliographystyle{ieee_fullname}
\bibliography{egbib}

\begin{thebibliography}{10}\itemsep=-1pt

\bibitem{an2012robust}
Lingling An, Xinbo Gao, Xuelong Li, Dacheng Tao, Cheng Deng, Jie Li, et~al.
\newblock Robust reversible watermarking via clustering and enhanced pixel-wise
  masking.
\newblock {\em IEEE Trans. Image Processing}, 21(8):3598--3611, 2012.

\bibitem{cai2009locality}
Deng Cai, Xiaofei He, Xuanhui Wang, Hujun Bao, and Jiawei Han.
\newblock Locality preserving nonnegative matrix factorization.
\newblock In {\em IJCAI}, volume~9, pages 1010--1015, 2009.

\bibitem{chen2018deep}
Pu Chen, Xinyi Xu, and Cheng Deng.
\newblock Deep view-aware metric learning for person re-identification.
\newblock In {\em IJCAI}, pages 620--626, 2018.

\bibitem{chen2011large}
Xinlei Chen and Deng Cai.
\newblock Large scale spectral clustering with landmark-based representation.
\newblock In {\em AAAI}, volume~5, page~14, 2011.

\bibitem{chen2016infogan}
Xi Chen, Yan Duan, Rein Houthooft, John Schulman, Ilya Sutskever, and Pieter
  Abbeel.
\newblock Infogan: Interpretable representation learning by information
  maximizing generative adversarial nets.
\newblock In {\em Advances in neural information processing systems}, pages
  2172--2180, 2016.

\bibitem{chi2015splitting}
Eric~C Chi and Kenneth Lange.
\newblock Splitting methods for convex clustering.
\newblock {\em Journal of Computational and Graphical Statistics},
  24(4):994--1013, 2015.

\bibitem{dalal2005histograms}
Navneet Dalal and Bill Triggs.
\newblock Histograms of oriented gradients for human detection.
\newblock In {\em Computer Vision and Pattern Recognition, 2005. CVPR 2005.
  IEEE Computer Society Conference on}, volume~1, pages 886--893. IEEE, 2005.

\bibitem{deng2019unsupervised}
C Deng, E Yang, T Liu, W Liu, J Li, and D Tao.
\newblock Unsupervised semantic-preserving adversarial hashing for image
  search.
\newblock {\em IEEE transactions on image processing: a publication of the IEEE
  Signal Processing Society}, 2019.

\bibitem{dilokthanakul2016deep}
Nat Dilokthanakul, Pedro~AM Mediano, Marta Garnelo, Matthew~CH Lee, Hugh
  Salimbeni, Kai Arulkumaran, and Murray Shanahan.
\newblock Deep unsupervised clustering with gaussian mixture variational
  autoencoders.
\newblock {\em arXiv preprint arXiv:1611.02648}, 2016.

\bibitem{dizaji2017deep}
Kamran~Ghasedi Dizaji, Amirhossein Herandi, Cheng Deng, Weidong Cai, and Heng
  Huang.
\newblock Deep clustering via joint convolutional autoencoder embedding and
  relative entropy minimization.
\newblock In {\em Computer Vision (ICCV), 2017 IEEE International Conference
  on}, pages 5747--5756. IEEE, 2017.

\bibitem{gu2009co}
Quanquan Gu and Jie Zhou.
\newblock Co-clustering on manifolds.
\newblock In {\em Proceedings of the 15th ACM SIGKDD international conference
  on Knowledge discovery and data mining}, pages 359--368. ACM, 2009.

\bibitem{guo2017improved}
Xifeng Guo, Long Gao, Xinwang Liu, and Jianping Yin.
\newblock Improved deep embedded clustering with local structure preservation.
\newblock In {\em International Joint Conference on Artificial Intelligence
  (IJCAI-17)}, pages 1753--1759, 2017.

\bibitem{hjelm2018learning}
R~Devon Hjelm, Alex Fedorov, Samuel Lavoie-Marchildon, Karan Grewal, Adam
  Trischler, and Yoshua Bengio.
\newblock Learning deep representations by mutual information estimation and
  maximization.
\newblock {\em arXiv preprint arXiv:1808.06670}, 2018.

\bibitem{hoi2010semi}
Steven~CH Hoi, Wei Liu, and Shih-Fu Chang.
\newblock Semi-supervised distance metric learning for collaborative image
  retrieval and clustering.
\newblock {\em ACM Transactions on Multimedia Computing, Communications, and
  Applications (TOMM)}, 6(3):18, 2010.

\bibitem{jaiswal2018unsupervised}
Ayush Jaiswal, Rex~Yue Wu, Wael Abd-Almageed, and Prem Natarajan.
\newblock Unsupervised adversarial invariance.
\newblock In {\em Advances in Neural Information Processing Systems}, pages
  5097--5107, 2018.

\bibitem{ji2017deep}
Pan Ji, Tong Zhang, Hongdong Li, Mathieu Salzmann, and Ian Reid.
\newblock Deep subspace clustering networks.
\newblock In {\em Advances in Neural Information Processing Systems}, pages
  24--33, 2017.

\bibitem{jiang2012transfer}
Wenhao Jiang and Fu-lai Chung.
\newblock Transfer spectral clustering.
\newblock In {\em Joint European Conference on Machine Learning and Knowledge
  Discovery in Databases}, pages 789--803. Springer, 2012.

\bibitem{jiang2016variational}
Zhuxi Jiang, Yin Zheng, Huachun Tan, Bangsheng Tang, and Hanning Zhou.
\newblock Variational deep embedding: An unsupervised and generative approach
  to clustering.
\newblock {\em arXiv preprint arXiv:1611.05148}, 2016.

\bibitem{kuhn1955hungarian}
Harold~W Kuhn.
\newblock The hungarian method for the assignment problem.
\newblock {\em Naval research logistics quarterly}, 2(1-2):83--97, 1955.

\bibitem{lecun1998gradient}
Yann LeCun, L{\'e}on Bottou, Yoshua Bengio, and Patrick Haffner.
\newblock Gradient-based learning applied to document recognition.
\newblock {\em Proceedings of the IEEE}, 86(11):2278--2324, 1998.

\bibitem{li2018self}
Chao Li, Cheng Deng, Ning Li, Wei Liu, Xinbo Gao, and Dacheng Tao.
\newblock Self-supervised adversarial hashing networks for cross-modal
  retrieval.
\newblock In {\em CVPR}, pages 4242--4251, 2018.

\bibitem{li2019coupled}
Chao Li, Cheng Deng, Lei Wang, De Xie, and Xianglong Liu.
\newblock Coupled cyclegan: Unsupervised hashing network for cross-modal
  retrieval.
\newblock {\em arXiv preprint arXiv:1903.02149}, 2019.

\bibitem{li2018discriminatively}
Fengfu Li, Hong Qiao, and Bo Zhang.
\newblock Discriminatively boosted image clustering with fully convolutional
  auto-encoders.
\newblock {\em Pattern Recognition}, 83:161--173, 2018.

\bibitem{li2016scalable}
Yeqing Li, Junzhou Huang, and Wei Liu.
\newblock Scalable sequential spectral clustering.
\newblock In {\em Thirtieth AAAI conference on artificial intelligence}, 2016.

\bibitem{liu2010large}
Wei Liu, Junfeng He, and Shih-Fu Chang.
\newblock Large graph construction for scalable semi-supervised learning.
\newblock In {\em Proceedings of the 27th international conference on machine
  learning (ICML-10)}, pages 679--686, 2010.

\bibitem{maas2013rectifier}
Andrew~L Maas, Awni~Y Hannun, and Andrew~Y Ng.
\newblock Rectifier nonlinearities improve neural network acoustic models.
\newblock In {\em Proc. icml}, volume~30, page~3, 2013.

\bibitem{maaten2008visualizing}
Laurens van~der Maaten and Geoffrey Hinton.
\newblock Visualizing data using t-sne.
\newblock {\em Journal of machine learning research}, 9(Nov):2579--2605, 2008.

\bibitem{macqueen1967some}
James MacQueen et~al.
\newblock Some methods for classification and analysis of multivariate
  observations.
\newblock In {\em Proceedings of the fifth Berkeley symposium on mathematical
  statistics and probability}, volume~1, pages 281--297. Oakland, CA, USA,
  1967.

\bibitem{masci2011stacked}
Jonathan Masci, Ueli Meier, Dan Cire{\c{s}}an, and J{\"u}rgen Schmidhuber.
\newblock Stacked convolutional auto-encoders for hierarchical feature
  extraction.
\newblock In {\em International Conference on Artificial Neural Networks},
  pages 52--59. Springer, 2011.

\bibitem{mukherjee2018clustergan}
Sudipto Mukherjee, Himanshu Asnani, Eugene Lin, and Sreeram Kannan.
\newblock Clustergan: Latent space clustering in generative adversarial
  networks.
\newblock {\em arXiv preprint arXiv:1809.03627}, 2018.

\bibitem{ng2002spectral}
Andrew~Y Ng, Michael~I Jordan, and Yair Weiss.
\newblock On spectral clustering: Analysis and an algorithm.
\newblock In {\em Advances in neural information processing systems}, pages
  849--856, 2002.

\bibitem{ng2003sift}
Pauline~C Ng and Steven Henikoff.
\newblock Sift: Predicting amino acid changes that affect protein function.
\newblock {\em Nucleic acids research}, 31(13):3812--3814, 2003.

\bibitem{nowozin2016f}
Sebastian Nowozin, Botond Cseke, and Ryota Tomioka.
\newblock f-gan: Training generative neural samplers using variational
  divergence minimization.
\newblock In {\em Advances in Neural Information Processing Systems}, pages
  271--279, 2016.

\bibitem{shah2017robust}
Sohil~Atul Shah and Vladlen Koltun.
\newblock Robust continuous clustering.
\newblock {\em Proceedings of the National Academy of Sciences},
  114(37):9814--9819, 2017.

\bibitem{shah2018deep}
Sohil~Atul Shah and Vladlen Koltun.
\newblock Deep continuous clustering.
\newblock {\em arXiv preprint arXiv:1803.01449}, 2018.

\bibitem{shaham2018spectralnet}
Uri Shaham, Kelly Stanton, Henry Li, Boaz Nadler, Ronen Basri, and Yuval
  Kluger.
\newblock Spectralnet: Spectral clustering using deep neural networks.
\newblock {\em arXiv preprint arXiv:1801.01587}, 2018.

\bibitem{shi2000normalized}
Jianbo Shi and Jitendra Malik.
\newblock Normalized cuts and image segmentation.
\newblock {\em IEEE Transactions on pattern analysis and machine intelligence},
  22(8):888--905, 2000.

\bibitem{tzoreff2018deep}
Elad Tzoreff, Olga Kogan, and Yoni Choukroun.
\newblock Deep discriminative latent space for clustering.
\newblock {\em arXiv preprint arXiv:1805.10795}, 2018.

\bibitem{xiao2017fashion}
Han Xiao, Kashif Rasul, and Roland Vollgraf.
\newblock Fashion-mnist: a novel image dataset for benchmarking machine
  learning algorithms.
\newblock {\em arXiv preprint arXiv:1708.07747}, 2017.

\bibitem{xie2016unsupervised}
Junyuan Xie, Ross Girshick, and Ali Farhadi.
\newblock Unsupervised deep embedding for clustering analysis.
\newblock In {\em International conference on machine learning}, pages
  478--487, 2016.

\bibitem{xu2003document}
Wei Xu, Xin Liu, and Yihong Gong.
\newblock Document clustering based on non-negative matrix factorization.
\newblock In {\em Proceedings of the 26th annual international ACM SIGIR
  conference on Research and development in informaion retrieval}, pages
  267--273. ACM, 2003.

\bibitem{yang2016towards}
Bo Yang, Xiao Fu, Nicholas~D Sidiropoulos, and Mingyi Hong.
\newblock Towards k-means-friendly spaces: Simultaneous deep learning and
  clustering.
\newblock {\em arXiv preprint arXiv:1610.04794}, 2016.

\bibitem{yang2018semantic}
Erkun Yang, Cheng Deng, Tongliang Liu, Wei Liu, and Dacheng Tao.
\newblock Semantic structure-based unsupervised deep hashing.
\newblock In {\em IJCAI}, pages 1064--1070, 2018.

\bibitem{yang2016joint}
Jianwei Yang, Devi Parikh, and Dhruv Batra.
\newblock Joint unsupervised learning of deep representations and image
  clusters.
\newblock In {\em Proceedings of the IEEE Conference on Computer Vision and
  Pattern Recognition}, pages 5147--5156, 2016.

\bibitem{yang2019weightreg}
Muli Yang, Cheng Deng, and Feiping Nie.
\newblock Adaptive-weighting discriminative regression for multi-view
  classification.
\newblock {\em Pattern Recogn.}, 88(4):236--245, 2019.

\bibitem{Yang2018NewL1}
Xu Yang, Cheng Deng, Xianglong Liu, and Feiping Nie.
\newblock New l2, 1-norm relaxation of multi-way graph cut for clustering.
\newblock In {\em AAAI}, 2018.

\bibitem{yi2015efficient}
Jinfeng Yi, Lijun Zhang, Tianbao Yang, Wei Liu, and Jun Wang.
\newblock An efficient semi-supervised clustering algorithm with sequential
  constraints.
\newblock In {\em Proceedings of the 21th ACM SIGKDD International Conference
  on Knowledge Discovery and Data Mining}, pages 1405--1414. ACM, 2015.

\bibitem{yu2018incremental}
Tianshu Yu, Junchi Yan, Wei Liu, and Baoxin Li.
\newblock Incremental multi-graph matching via diversity and randomness based
  graph clustering.
\newblock In {\em Proceedings of the European Conference on Computer Vision
  (ECCV)}, pages 139--154, 2018.

\bibitem{zhang2013agglomerative}
Wei Zhang, Deli Zhao, and Xiaogang Wang.
\newblock Agglomerative clustering via maximum incremental path integral.
\newblock {\em Pattern Recognition}, 46(11):3056--3065, 2013.

\bibitem{zhou2018deep}
Pan Zhou, Yunqing Hou, and Jiashi Feng.
\newblock Deep adversarial subspace clustering.
\newblock In {\em Proceedings of the IEEE Conference on Computer Vision and
  Pattern Recognition}, pages 1596--1604, 2018.

\end{thebibliography}
}

\end{document}